\pdfoutput=1

\documentclass[11pt]{article}

\usepackage[preprint]{acl}

\usepackage{times}
\usepackage{latexsym}

\usepackage[T1]{fontenc}

\usepackage[utf8]{inputenc}

\usepackage{microtype}

\usepackage{inconsolata}

\usepackage{graphicx}

\def\thickhline{\noalign{\hrule height 1.2pt}}

\usepackage{fancyvrb}
\usepackage{amsmath}

\title{Surprisal from Larger Transformer-based Language Models\\Predicts fMRI Data More Poorly}

\author{Yi-Chien Lin \\
  The Ohio State University \\
  Department of Linguistics \\
  \texttt{lin.4434@osu.edu} \\\And
  William Schuler \\
  The Ohio State University \\
  Department of Linguistics \\
  \texttt{schuler.77@osu.edu} \\}

\begin{document}
\maketitle
\begin{abstract}
There has been considerable interest in using surprisal from Transformer-based language models (LMs) as predictors of human sentence processing difficulty.
Recent work has observed an inverse scaling relationship between Transformers' per-word estimated probability and the predictive power of their surprisal estimates on reading times, showing that LMs with more parameters and trained on more data are {\em less} predictive of human reading times.
However, these studies focused on predicting latency-based measures.
Tests on brain imaging data have not shown a trend in any direction when using a relatively small set of LMs, leaving open the possibility that the inverse scaling phenomenon is constrained to latency data.
This study therefore conducted a more comprehensive evaluation using surprisal estimates from 17 pre-trained LMs across three different LM families on two functional magnetic resonance imaging (fMRI) datasets.
Results show that the inverse scaling relationship between models' per-word estimated probability and model fit on both datasets still obtains, resolving the inconclusive results of previous work and indicating that this trend is not specific to latency-based measures.

\end{abstract}

\section{Introduction}

Expectation-based theories of syntactic comprehension \citep{hale01,levy08} posit that comprehenders evaluate multiple hypotheses of a structure in parallel.
Within this line of research, processing difficulty can be quantified using information-theoretic surprisal \citep{shannon48} and the processing difficulty of a word is proportional to its surprisal estimate (i.e., negative log probability given the preceding context).

As Transformer-based models \citep{vaswanietal17} become more widely incorporated into natural language processing tasks, there has been considerable interest in using surprisal from these models as predictors of processing difficulty \citep{merkxfrank21,wilcoxetal20}.
Recent work has observed that surprisal from larger Transformer-based LMs, which have more parameters and lower model perplexity, is less predictive of reading times \citep{oh2022comparison,oh2023does,de-varda-marelli-2023-scaling,shainetal2024,oh-etal-2024-frequency}.
Larger LMs trained on larger amounts of data become more predictive of rare tokens throughout the course of training, resulting in lower surprisal estimates that diverge from human reading times \citep{oh-etal-2024-frequency}.
In addition to reading times, \citet{oh2022comparison} also use surprisal estimates from Transformers to predict brain imaging data, but observe no inverse or positive scaling when using a relatively small set of LMs.
However, moving away from surprisal, recent studies observed a positive scaling on brain imaging data when using vectors directly from large language models as predictors \citep{schrimpf2021neural,hosseini2024artificial} so it may be tempting to think brain imaging data behaves differently from latency-based data (i.e., reading times).

This study presents evidence against this possibility by evaluating the scaling trend of the predictive power of surprisal estimates from a comprehensive set of 17 Transformer-based LMs across three LM families on two different brain imaging datasets: the original dataset \citep{shain2020fmri}, which was also evaluated in \citet{oh2022comparison}, and a separate dataset \citep{pereira2018}.
Results show statistically significant inverse scalings on both datasets which resolve the inconclusive results from \citet{oh2022comparison} on \citet{shain2020fmri} when using surprisal estimates from a larger set of LMs and replicate inverse scaling on \citet{pereira2018}.

This work provides an evidence showing that the inverse scaling of surprisal estimates is not specific to latency-based data.
This piece of evidence provides a more complete picture of the relationship between the model sizes and predictive power of surprisal estimates on psychometric data, which could potentially be helpful in providing researchers with insights into the appropriate use of LMs in understanding human sentence processing.

\section{Related Work}

Previous studies have examined the predictive power of surprisal from various types of LMs such as \textit{n}-gram, Long Short-Term Memory network \citep{hochreiterschmidhuber97}, and recurrent neural network models on psychometric data (e.g., \citealp{goodkindbicknell18}; \citealp{wilcoxetal20}).
These studies observed a consistent trend in the positive relationship between a model's quality and its predictive power on psychometric data such as reading times: the better an LM's quality (or the higher an LM's per-word estimated probability), the better its fit to the psychometric data.

Recently, there has been increasing interest in comparing surprisal estimates from Transformer-based models against behavioral and neural measures of processing difficulty.
Differing from previous findings, \citet{oh2022comparison} observed a completely opposite {\em inverse} scaling relationship between LMs' per-word estimated probability and models' fit to reading times when using surprisal estimates from a set of pre-trained GPT-2 models \citep{radford2019language}.
In other words, larger LMs (with higher per-word estimated probability) are less predictive of reading times.
This observation was further replicated using surprisal estimates from a larger set of Transformer-based LMs \citep{oh2023does,oh-etal-2024-frequency}.

Moreover, recent work in this line of study found that surprisal estimates from neural LMs have a tendency to underpredict reading times.
\citet{van2021single} and \citet{arehalli-etal-2022-syntactic} showed that neural LM surprisal successfully predicts human precessing difficulty in garden-path constructions but consistently underestimates its magnitude.
\citet{kuribayashi-etal-2022-context} also observed that surprisal estimates from neural LMs without context limitations underpredict reading times of English and Japanese naturalistic text.
Furthermore, \citet{oh2023does} conducted an analysis of the inverse scaling relationship between Transformer-based LMs' per-word estimated probability and the predictive power of surprisal estimates, observing that the poorer fit to reading times of larger LMs is mainly driven by the lower surprisal values those LMs assign to open-class words such as nouns and adjectives.
\citet{oh-etal-2024-frequency} followed up on this observation and showed that word frequency explains the inverse correlation between the size of Transformer-based LMs, the amount of training data, and the predictive power of surprisal estimates.
They found that larger LMs get better at predicting rare words, resulting in lower surprisal which diverges from human reading times.

\section{Experiment 1: Predictive Power of Surprisal on \citet{shain2020fmri}}\label{exp1}

Experiment~1 revisited \citet{oh2022comparison} and evaluated the predictive power of surprisal estimates from a larger set of Transformer-based LMs on \citet{shain2020fmri} (hereafter `\textbf{Natural Stories fMRI}').
We collected surprisal estimates from 17 pre-trained autoregressive Transformer-based LMs and used them to predict the blood oxygenation level-dependent (BOLD) signals from Natural Stories fMRI.
The experiment setup followed that of \citet{shain2020fmri}.\footnote{Code for replicating the experiments in this study is available at \url{https://github.com/modelblocks/modelblocks-release}.}


\subsection{Response Data}\label{exp1-resp-data}

Natural Stories fMRI \citep{shain2020fmri} was collected at a two-second fixed time interval from 78 subjects while they listen to a recording of the Natural Stories Corpus \citep{futrell2021natural} which consists of naturalistic English stories.
This dataset includes the time series of BOLD signals of several functional regions of interest (fROIs) in the language network \citep{fedorenko2011functional}, which were identified with a localizer task.
For each fROI, the BOLD signals were averaged across all voxels within that fROI.

Following \citet{shain2020fmri}, Natural Stories fMRI was partitioned into fit, exploratory, and held-out sets which contain around 50\%, 25\%, and 25\% of the data points, respectively.
Specifically, we grouped the BOLD signals into 30-second chunks and distributed those chunks into the sets to ensure the time series data in the sets are sequential and to avoid having data which is temporally too close to each other in each set. 
This resulted in 100,084, 50,818, and 51,393 data points in the fit, exploratory, and held-out sets.
The regression models were fit using the fit set and the results reported in this study are on the held-out set.

\subsection{Predictors}\label{exp1-predictors}

We collected surprisal estimates from model variants of three autoregressive Transformer-based LM families which were used in \citet{oh2022comparison} and \citet{oh2023does}: GPT-2 \citep{radford2019language}, GPT-Neo \citep{blacketal21,black-etal-2022-gpt,gpt-j}, and OPT \citep{zhang2022opt}\footnote{Due to the constraints in computational resources available to us, we did not include the largest OPT model variant, which has around 175 billion parameters.} families.
The hyperparameters of each LM are detailed in Table~\ref{table:hyper-para}, Appendix~\ref{appendix-a}.

Each story of the Natural Stories Corpus was first tokenized using the byte-pair encoding \citep{sennrich-etal-2016-neural} tokenizer corresponding to each model variant.
The tokenized texts were then input to each LM to calculate token-wise surprisal estimates.
In cases where a tokenized story exceeds a single context window of the LMs, to calculate the surprisal estimates for the remaining tokens, the latter half of the previous context window was used as the first half of the context window.
If a word consists of multiple subword tokens, we aggregate the token-level surprisal estimates to get the word-level surprisal estimate.\footnote{Although we tokenized the text at token-level instead of word-level, the relationship between the predictive power of the surprisal and the LM quality is unlikely to make much difference, as \citet{oh-schuler-2025-impact} found that there is a relatively small effect of token granularity on the relationship.}

Recent studies \citep{oh-schuler-2024-leading,pimentel-meister-2024-compute} have shown that, for languages which use whitespace characters in their orthography, the leading whitespaces (i.e., the whitespace characters which immediately precede the tokens) added by LMs' tokenizers can result in a sum over all word probabilities that is greater than one.
This study therefore incorporated whitespace-trailing decoding into surprisal calculation.
For each word, the probability of the leading whitespace was re-allocated to its preceding word when calculating word-level probabilities.

Since the response data of Natural Stories fMRI is the time series of BOLD signals, we first collected the surprisal estimates of the words and then applied a hemodynamic response function (HRF; \citealt{boynton1996linear}) to convolve those surprisal estimates so that the convolved surprisal estimates align with the response data.
A canonical HRF convolves predictors into the shape of the BOLD signals when a subject is presented with a stimulus.
The HRF we used for this experiment is shown in the following equation, giving magnitude $f(x)$ for offset time $x$ in seconds:

\begin{small}
{\begin{eqnarray}
    f(x) = \frac{5.2^{5.4} x^{5.4 - 1} e^{-5.2 x}}{\Gamma(5.4)} - 0.35 \, \frac{7.35^{10.8} x^{10.8 - 1} e^{-7.35 x}}{\Gamma(10.8)} \nonumber
    \label{eq:hrf}
\end{eqnarray}}
\end{small}


\subsection{Regression Modeling}\label{reg-modeling}

Once the surprisal estimates were collected, we fit a set of linear mixed-effect (LME) models to BOLD signals with \texttt{lme4} \citep{bates2015fitting} using the fit set.
During regression modeling, in addition to surprisal estimates, we also included a set of baseline predictors, including the index of the word position in the sentence and the word length in characters.
All models were fit using raw BOLD values along with the above baseline predictors.
The predictors were centered, scaled, and convolved with the HRF before being used to fit the LME models.
Following \citet{barr2013random}, during model fitting, we started with the maximal LME models and eliminated one random effect at a time until the LME models converge.
The maximal converging formula for Natural Stories fMRI was:

\begin{small}
\begin{verbatim}
BOLD ~ z.(sentpos) + z.(wlen) + z.(surp) + 
       (1 + z.(wlen) + z.(surp) | subject) + 
       (1 + z.(surp) | storyid)
\end{verbatim}
\end{small}
where \texttt{sentpos} is the word position in the sentence, \texttt{wlen} is word length, \texttt{surp} is surprisal estimate, and \texttt{storyid} is the index of a story.\footnote{The original design of the LME model for Natural Stories fMRI is shown in Appendix~\ref{original-max-lme}.}
We show the predictive power with the model log likelihood calculated on the held-out set.


\begin{figure}[t!]
    \centering
    \includegraphics[width=\columnwidth]{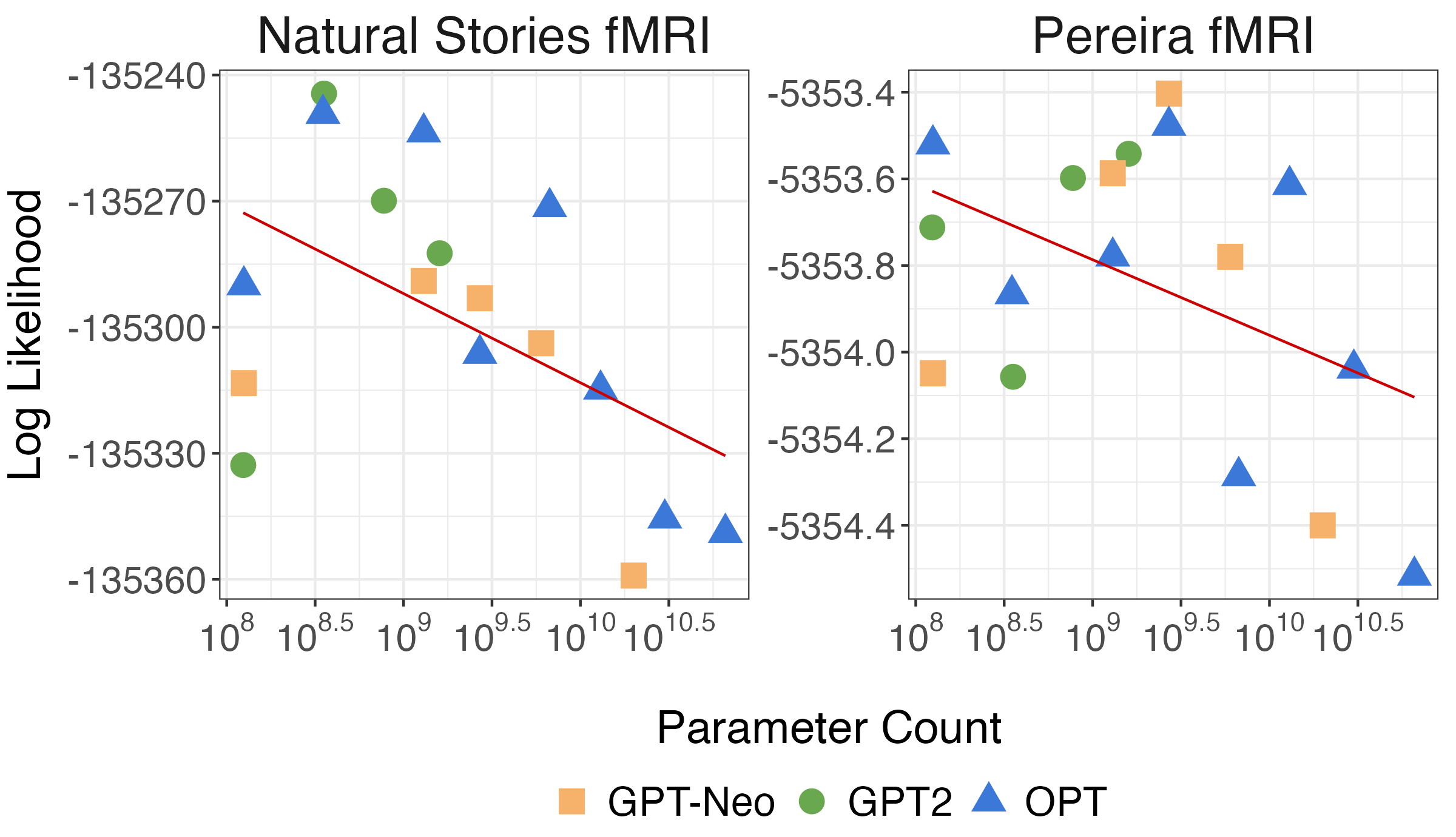}
    \caption{Predictive power of surprisal estimates from pretrained LMs on BOLD signals of Natural Stories fMRI (left) and Pereira fMRI (right).}
    \label{fig:results}
\end{figure}

\subsection{Results}

Figure \ref{fig:results} (left) presents the relationship between the regression model log likelihood and log-transformed parameter count for Natural Stories fMRI.
The results show a significant decrease in predictive power when parameter count (model size) increases ($p$~<~0.01 by a permutation test with 1,000 permutations), resolving the inconclusive results in \citet{oh2022comparison}, which only evaluate the predictive power of surprisal estimates from a relatively small set of LMs on this same dataset.\footnote{The choice of model families was based on previous work which observed inverse scaling on reading times \citep{oh2022comparison,oh2023does}. Based on the inverse scaling trend observed for Pythia and the similarity to inverse scaling results for larger off-the-shelf LLMs found in \citet{oh2023does,oh-etal-2024-frequency}, we expect more recent, larger models to yield substantially worse fits.}

\section{Experiment 2: Predictive Power of Surprisal on \citet{pereira2018}}\label{exp2}
Experiment~1 resolves the inconclusive results on Natural Stories fMRI from \citet{oh2022comparison} by showing an inverse scaling phenomenon with a larger set of LMs.
To examine whether this phenomenon is generalizable across other brain imaging datasets, we additionally evaluated the predictive power of the same set of LMs on a different fMRI dataset, \citet{pereira2018} (hereafter `\textbf{Pereira fMRI}').

\subsection{Response Data}
Pereira fMRI contains data from Experiments~II and III in \citet{pereira2018}.
Eight subjects participated in Experiment~II and read 96 English passages, which includes 384 sentences in total; six subjects participated in Experiment~III and read 72 English passages, which consists of 243 sentences in total.
Each passage consists of three to four sentences and sentences were presented to subjects one at a time.
Like Natural Stories fMRI, Pereira fMRI also includes BOLD signals from the voxels in the language network.
The responses of this dataset are per-sentence BOLD signals, which were obtained by aggregating activation across regions.
Following \citet{oh2022comparison}, Pereira fMRI was partitioned into fit, exploratory, and held-out sets, resulting in the sizes of 2,268, 1,132, and 1,130 data points respectively in each set.

\subsection{Procedure}
In general, the procedure of Experiment~2 followed that of Experiment~1 except that since the response data for Pereira fMRI is per-sentence BOLD signals, we followed recent work on the same dataset \citep{hosseini2024artificial} and used the surprisal estimate of the sentence-final word as the predictor for each response data point.\footnote{\citet{hosseini2024artificial} use the model representations of the sentence-final words and not the surprisal estimates as the representation of the entire sentence.}
The motivation of using the sentence-final word surprisal is that its preceding context has already been taken into consideration when calculating the surprisal estimates of the last words of the sentences.

When fitting the LME models, we used sentence length in words and index of the passage containing the sentence as baseline predictors.
Similar to the procedure in Experiment~1, following \citet{barr2013random}, we started with the maximal LME models and eliminated the random effects one by one until the models no longer have serious convergence issues. 
The maximal converging formula was:

\begin{small}
\begin{verbatim}
BOLD ~ z.(sentlen) + z.(passpos)+ z.(lastwordsurp) + 
       (1 + z.(sentlen) + z.(passpos) + 
        z.(lastwordsurp) | subject) + (1 + z.(sentlen) + 
        z.(lastwordsurp) | sentid)
\end{verbatim}
\end{small}
where \texttt{sentlen} is sentence length in words, \texttt{passpos} is the sentence position within passage, \texttt{lastwordsurp} is last-word surprisal estimate, and \texttt{sentid} is the index of each sentence.\footnote{The original design of the LME model for Pereira fMRI is shown in Appendix~\ref{original-max-lme}.}

\subsection{Results}
Figure~\ref{fig:results} (right) shows the results of Experiment~2.
Similar to the results of Experiment~1, the results on Pereira fMRI also show that as parameter count (model size) increases, the regression model log likelihood decreases significantly ($p$~<~0.05).
The replication of the inverse scaling phenomenon on a different dataset suggests that this phenomenon is not dataset-specific.\footnote{For both corpora, in addition to running experiments using LME models, per reviewers' request, we also ran linear regression for both datasets (1) using one data point per sentence and (2) averaging across participants.

For Pereira fMRI, since the response data is already per-sentence BOLD signals, for each sentence, we averaged across the BOLD signals across all participants. Similar to our method described in Experiment~2, which was based on the methodology in described \citet{hosseini2024artificial}, we used the last-word surprisal estimate as the representation of each data point. We then performed linear regression modeling using six-fold cross-validation. The test fold sizes range from 87 to 122 data points and the training fold sizes range from 505 to 540 data points.

For Natural Stories fMRI, since the response data is the time series of BOLD signals,
we do not have direct information about sentence segmentation. In order to obtain an estimate of the sentence-level BOLD signals and surprisal estimates, in general, we followed \citet{feghhi2024large} which used an fMRI dataset collected at a two-second time interval \citep{blank2014functional}, similar to Natural Stories fMRI. We mapped each response data point to the word (along with its surprisal estimate) that is closest to four seconds
before that data point. Note that for this analysis, we did not apply HRF convolution to the predictors. This allowed us to obtain an estimate of sentence segmentation points since we have the sentence id information of each word. To get the sentence-level BOLD signals, we averaged all BOLD signals that were mapped to the same sentence id. As for sentence-level surprisal estimates, it was not possible to use last-word surprisal since the BOLD signals are not sentence-aligned, so for each sentence, we averaged all surprisal estimates that were within that sentence and were mapped to some BOLD signals.

Once we obtained the sentence-level BOLD signals for each participant and surprisal estimates, we averaged across participants' responses. Since the number of data points were reduced to a small amount after obtaining the sentence-level responses/surprisal estimates and averaging across subjects, we performed linear regression modeling using eight-fold cross-validation. For each iteration, we used the data of one story as the test data and the remaining seven stories as the training data. The test fold sizes range from 210 to 390 data points and training fold sizes range from 1992 to 2172 data points.

For both corpora in this analysis, we report the mean squared errors averaged across each iteration. In this analyses, Natural Stories fMRI still shows a significant inverse scaling (p~<~0.05) while Pereira fMRI shows numerically but not statistically significant inverse scaling. We speculate that the reason why statistical significance was not observed for inverse scaling on Pereira fMRI under this experimental setting might be because the BOLD signals used for regression modeling were averaged across subjects, which could potentially make the data somewhat sparser and therefore underpowered.

}

\section{Conclusion}

Recent studies \citep{oh2022comparison,oh2023does,oh-etal-2024-frequency} have observed a robust negative relationship between Transformer-based LMs' quality and the predictive power of surprisal estimates from those LMs on latency data.
This work generalizes that result by examining the predictive power of surprisal estimates from Transformer-based models on brain imaging data.
The predictive power of surprisal estimates from 17 pre-trained LMs across three LM families was evaluated on two fMRI datasets. 
Results show that the inverse scaling between model size and model fit still obtains on brain imaging data.
The inverse relationship between the predictive power of surprisal estimates from those LMs and parameter count (model size) on brain imaging data is not as strict as that on latency-based data, but this may be due to the fact that brain imaging data is much noisier than latency-based data.
Nevertheless, the observation of the inverse scaling phenomenon on two distinct brain imaging datasets indicates that this trend is not specific to latency-based measures, suggesting that brain imaging data do not behave substantially differently from latency data.
We speculate the deviation of larger LMs from brain imaging data is due to the same reason for reading times \citep{oh2023does}.
This additional piece of evidence reinforces the observation of inverse scaling phenomenon on latency data \citep{oh2022comparison,oh2023does}, suggesting that smaller models could potentially be helpful in providing researchers with insights into the appropriate use of LMs in understanding sentence processing. 

\section*{Limitations}

This study attempts to examine the predictive power of surprisal estimates from Transformer-based models on fMRI data.
Large language models evaluated in this work are trained on English text and the datasets evaluated are collected from English speakers.
Therefore, these findings may or may not be replicated in other languages.

\section*{Ethics Statement}

The fMRI datasets used in this study are from published work \citep{shain2020fmri,pereira2018}.
Details regarding the data collection, validation and other relevant procedures are described in respective publications.
Since this work focuses on evaluating the predictive power of large language models on fMRI data, the potential risks and harmful impacts posed by this study on the society appear to be minimal.

\section*{Acknowledgments}

We thank the anonymous reviewers for their helpful feedback. This work was supported by the National Science Foundation grant \#2313140. All views expressed are those of the authors and do not necessarily reflect the views of the National Science Foundation.

\bibliography{custom}

@article{oh2022comparison,
  title={Comparison of structural parsers and neural language models as surprisal estimators},
  author={Oh, Byung-Doh and Clark, Christian and Schuler, William},
  journal={Frontiers in Artificial Intelligence},
  volume={5},
  pages={777963},
  year={2022},
  publisher={Frontiers},
  url={https://doi.org/10.3389/frai.2022.777963}
}

@article{oh2023does,
    author = {Oh, Byung-Doh and Schuler, William},
    title = {Why Does Surprisal From Larger {T}ransformer-Based Language Models Provide a Poorer Fit to Human Reading Times?},
    journal = {Transactions of the Association for Computational Linguistics},
    volume = {11},
    pages = {336-350},
    year = {2023},
    month = {03},
    issn = {2307-387X},
    doi = {10.1162/tacl_a_00548},
    url = {https://doi.org/10.1162/tacl\_a\_00548},
    eprint = {https://direct.mit.edu/tacl/article-pdf/doi/10.1162/tacl\_a\_00548/2075940/tacl\_a\_00548.pdf},
}

@article{radford2019language,
  title={Language models are unsupervised multitask learners},
  author={Radford, Alec and Wu, Jeffrey and Child, Rewon and Luan, David and Amodei, Dario and Sutskever, Ilya},
  journal={OpenAI Technical Report},
  year={2019},
  url = {https://cdn.openai.com/better-language-models/language_models_are_unsupervised_multitask_learners.pdf}
}

@article{blacketal21,
  author       = {Black, Sid and
                  Gao, Leo and
                  Wang, Phil and
                  Leahy, Connor and
                  Biderman, Stella},
  title        = {{GPT-Neo: L}arge Scale Autoregressive Language Modeling with {Mesh-Tensorflow}},
  month        = mar,
  year         = 2021,
  journal    = {Zenodo},
  version      = {1.0},
  doi          = {10.5281/zenodo.5297715},
}

@inproceedings{black-etal-2022-gpt,
    title = "{GPT}-{N}eo{X}-20{B}: An Open-Source Autoregressive Language Model",
    author = "Black, Sidney  and
      Biderman, Stella  and
      Hallahan, Eric  and
      Anthony, Quentin  and
      Gao, Leo  and
      Golding, Laurence  and
      He, Horace  and
      Leahy, Connor  and
      McDonell, Kyle  and
      Phang, Jason  and
      Pieler, Michael  and
      Prashanth, Usvsn Sai  and
      Purohit, Shivanshu  and
      Reynolds, Laria  and
      Tow, Jonathan  and
      Wang, Ben  and
      Weinbach, Samuel",
    editor = "Fan, Angela  and
      Ilic, Suzana  and
      Wolf, Thomas  and
      Gall{\'e}, Matthias",
    booktitle = "Proceedings of BigScience Episode {\#}5 -- Workshop on Challenges {\&} Perspectives in Creating Large Language Models",
    month = may,
    year = "2022",
    address = "virtual+Dublin",
    publisher = "Association for Computational Linguistics",
    url = "https://aclanthology.org/2022.bigscience-1.9",
    doi = "10.18653/v1/2022.bigscience-1.9",
    pages = "95--136",
}

@misc{gpt-j,
  author = {Wang, Ben and Komatsuzaki, Aran},
  title = {{GPT-J-6B: A 6 Billion Parameter Autoregressive Language Model}},
  howpublished = {\url{https://github.com/kingoflolz/mesh-transformer-jax}},
  year = 2021,
  month = May
}

@misc{zhang2022opt,
      title={{OPT}: Open Pre-trained Transformer Language Models}, 
      author={Susan Zhang and Stephen Roller and Naman Goyal and Mikel Artetxe and Moya Chen and Shuohui Chen and Christopher Dewan and Mona Diab and Xian Li and Xi Victoria Lin and Todor Mihaylov and Myle Ott and Sam Shleifer and Kurt Shuster and Daniel Simig and Punit Singh Koura and Anjali Sridhar and Tianlu Wang and Luke Zettlemoyer},
      year={2022},
      eprint={2205.01068},
      archivePrefix={arXiv},
      primaryClass={cs.CL}
}

@inproceedings{hale01,
    title = "A Probabilistic {E}arley Parser as a Psycholinguistic Model",
    author = "Hale, John",
    booktitle = "Second Meeting of the North {A}merican Chapter of the Association for Computational Linguistics",
    year = "2001",
    url = "https://aclanthology.org/N01-1021",
}

@article{levy08,
title = {Expectation-based syntactic comprehension},
journal = {Cognition},
volume = {106},
number = {3},
pages = {1126-1177},
year = {2008},
issn = {0010-0277},
doi = {https://doi.org/10.1016/j.cognition.2007.05.006},
url = {https://www.sciencedirect.com/science/article/pii/S0010027707001436},
author = {Roger Levy},
}

@article{shannon48,
  author = {{Shannon}, Claude},
  journal = {Bell System Technical Journal},
  pages = {379--423, 623--656},
  title = {A Mathematical Theory of Communication},
  volume = {27},
  year = {1948},
  doi={https://doi.org/10.1002/j.1538-7305.1948.tb01338.x}
}

@article{pereira2018, 
    title={Toward a universal decoder of linguistic meaning from brain activation}, 
    volume={9}, 
    ISSN={2041-1723}, 
    DOI={10.1038/s41467-018-03068-4}, 
    number={1},
    journal={Nature Communications},
    author={Pereira, Francisco and Lou, Bin and Pritchett, Brianna and Ritter, Samuel and Gershman, Samuel J. and Kanwisher, Nancy and Botvinick, Matthew and Fedorenko, Evelina},
    year={2018}, 
    month=mar, 
    pages={963}, 
    language={en}
}

@article{shain2020fmri,
  title={f{MRI} reveals language-specific predictive coding during naturalistic sentence comprehension},
  author={Shain, Cory and Blank, Idan Asher and van Schijndel, Marten and Schuler, William and Fedorenko, Evelina},
  journal={Neuropsychologia},
  volume={138},
  pages={107307},
  year={2020},
  publisher={Elsevier},
  doi={https://doi.org/10.1016/j.neuropsychologia.2019.107307}
}

@article{bates2015fitting,
  title={Fitting linear mixed-effects models using lme4},
  author={Bates, Douglas and M{\"a}chler, Martin and Bolker, Ben and Walker, Steve},
  journal={Journal of statistical software},
  volume={67},
  pages={1--48},
  year={2015},
  doi={https://doi.org/10.18637/jss.v067.i01}
}

@article{fedorenko2011functional,
  title={Functional specificity for high-level linguistic processing in the human brain},
  author={Fedorenko, Evelina and Behr, Michael K. and Kanwisher, Nancy},
  journal={Proceedings of the National Academy of Sciences},
  volume={108},
  number={39},
  pages={16428--16433},
  year={2011},
  publisher={National Academy of Sciences},
  doi={https://doi.org/10.1073/pnas.1112937108}
}

@inproceedings{vaswanietal17,
 author = {Vaswani, Ashish and Shazeer, Noam and Parmar, Niki and Uszkoreit, Jakob and Jones, Llion and Gomez, Aidan N and Kaiser, {\L}ukasz and Polosukhin, Illia},
 booktitle = {Advances in Neural Information Processing Systems},
 editor = {I. Guyon and U. Von Luxburg and S. Bengio and H. Wallach and R. Fergus and S. Vishwanathan and R. Garnett},
 pages = {},
 publisher = {Curran Associates, Inc.},
 title = {Attention is All you Need},
 url = {https://proceedings.neurips.cc/paper_files/paper/2017/file/3f5ee243547dee91fbd053c1c4a845aa-Paper.pdf},
 volume = {30},
 year = {2017}
}

@inproceedings{merkxfrank21,
    title = "Human Sentence Processing: Recurrence or Attention?",
    author = "Merkx, Danny  and
      Frank, Stefan L.",
    booktitle = "Proceedings of the Workshop on Cognitive Modeling and Computational Linguistics",
    month = jun,
    year = "2021",
    url = "https://aclanthology.org/2021.cmcl-1.2",
    doi = "10.18653/v1/2021.cmcl-1.2",
    pages = "12--22",
}

@inproceedings{wilcoxetal20,
    title = "On the Predictive Power of Neural Language Models for Human Real-Time Comprehension Behavior",
    author = "Ethan Gotlieb Wilcox and
    Jon Gauthier and
    Jennifer Hu and
    Peng Qian and
    Roger P. Levy",
    booktitle = "Proceedings of the 42nd Annual Meeting of the Cognitive Science Society",
    year = "2020",
    url = "https://cognitivesciencesociety.org/cogsci20/papers/0375",
    pages = "1707--1713"
}

@article{shainetal2024,
author = {Cory Shain  and Clara Meister  and Tiago Pimentel  and Ryan Cotterell  and Roger Levy },
title = {Large-scale evidence for logarithmic effects of word predictability on reading time},
journal = {Proceedings of the National Academy of Sciences},
volume = {121},
number = {10},
pages = {e2307876121},
year = {2024},
doi = {10.1073/pnas.2307876121},
URL = {https://www.pnas.org/doi/abs/10.1073/pnas.2307876121},
eprint = {https://www.pnas.org/doi/pdf/10.1073/pnas.2307876121},
}

@inproceedings{de-varda-marelli-2023-scaling,
    title = "Scaling in Cognitive Modelling: a Multilingual Approach to Human Reading Times",
    author = "de Varda, Andrea  and
      Marelli, Marco",
    editor = "Rogers, Anna  and
      Boyd-Graber, Jordan  and
      Okazaki, Naoaki",
    booktitle = "Proceedings of the 61st Annual Meeting of the Association for Computational Linguistics (Volume 2: Short Papers)",
    month = jul,
    year = "2023",
    address = "Toronto, Canada",
    publisher = "Association for Computational Linguistics",
    url = "https://aclanthology.org/2023.acl-short.14/",
    doi = "10.18653/v1/2023.acl-short.14",
    pages = "139--149"
}

@inproceedings{oh-etal-2024-frequency,
    title = "Frequency Explains the Inverse Correlation of Large Language Models' Size, Training Data Amount, and Surprisal's Fit to Reading Times",
    author = "Oh, Byung-Doh  and
      Yue, Shisen  and
      Schuler, William",
    editor = "Graham, Yvette  and
      Purver, Matthew",
    booktitle = "Proceedings of the 18th Conference of the European Chapter of the Association for Computational Linguistics (Volume 1: Long Papers)",
    month = mar,
    year = "2024",
    address = "St. Julian{'}s, Malta",
    publisher = "Association for Computational Linguistics",
    url = "https://aclanthology.org/2024.eacl-long.162/",
    pages = "2644--2663"
}

@article{hochreiterschmidhuber97,
  author      = {Sepp Hochreiter and Jürgen Schmidhuber},
  journal     = {Neural Computation},
  title       = {Long Short-Term Memory},
  year        = {1997},
  number      = {8},
  pages       = {1735--1780},
  volume      = {9},
  doi      = {10.1162/neco.1997.9.8.1735}
}

@inproceedings{goodkindbicknell18,
    title = "Predictive power of word surprisal for reading times is a linear function of language model quality",
    author = "Goodkind, Adam  and
      Bicknell, Klinton",
    editor = "Sayeed, Asad  and
      Jacobs, Cassandra  and
      Linzen, Tal  and
      van Schijndel, Marten",
    booktitle = "Proceedings of the 8th Workshop on Cognitive Modeling and Computational Linguistics ({CMCL} 2018)",
    month = jan,
    year = "2018",
    address = "Salt Lake City, Utah",
    publisher = "Association for Computational Linguistics",
    url = "https://aclanthology.org/W18-0102",
    doi = "10.18653/v1/W18-0102",
    pages = "10--18",
}

@article{van2021single,
  title={Single-stage prediction models do not explain the magnitude of syntactic disambiguation difficulty},
  author={Van Schijndel, Marten and Linzen, Tal},
  journal={Cognitive science},
  volume={45},
  number={6},
  pages={e12988},
  year={2021},
  publisher={Wiley Online Library},
  doi={10.1111/cogs.12988}
}

@inproceedings{arehalli-etal-2022-syntactic,
    title = "Syntactic Surprisal From Neural Models Predicts, But Underestimates, Human Processing Difficulty From Syntactic Ambiguities",
    author = "Arehalli, Suhas  and
      Dillon, Brian  and
      Linzen, Tal",
    booktitle = "Proceedings of the 26th Conference on Computational Natural Language Learning (CoNLL)",
    month = dec,
    year = "2022",
    address = "Abu Dhabi, United Arab Emirates (Hybrid)",
    publisher = "Association for Computational Linguistics",
    url = "https://aclanthology.org/2022.conll-1.20/",
    doi = "10.18653/v1/2022.conll-1.20",
    pages = "301--313",
}

@inproceedings{kuribayashi-etal-2022-context,
    title = "Context Limitations Make Neural Language Models More Human-Like",
    author = "Kuribayashi, Tatsuki  and
      Oseki, Yohei  and
      Brassard, Ana  and
      Inui, Kentaro",
    editor = "Goldberg, Yoav  and
      Kozareva, Zornitsa  and
      Zhang, Yue",
    booktitle = "Proceedings of the 2022 Conference on Empirical Methods in Natural Language Processing",
    month = dec,
    year = "2022",
    address = "Abu Dhabi, United Arab Emirates",
    publisher = "Association for Computational Linguistics",
    url = "https://aclanthology.org/2022.emnlp-main.712/",
    doi = "10.18653/v1/2022.emnlp-main.712",
    pages = "10421--10436"
}

@article{futrell2021natural,
  title={The Natural Stories corpus: a reading-time corpus of English texts containing rare syntactic constructions},
  author={Futrell, Richard and Gibson, Edward and Tily, Harry J. and Blank, Idan and Vishnevetsky, Anastasia and Piantadosi, Steven T. and Fedorenko, Evelina},
  journal={Language Resources and Evaluation},
  volume={55},
  pages={63--77},
  year={2021},
  publisher={Springer},
  url={https://doi.org/10.1007/s10579-020-09503-7},
  doi={10.1007/s10579-020-09503-7}
}

@inproceedings{sennrich-etal-2016-neural,
    title = "Neural Machine Translation of Rare Words with Subword Units",
    author = "Sennrich, Rico  and
      Haddow, Barry  and
      Birch, Alexandra",
    booktitle = "Proceedings of the 54th Annual Meeting of the Association for Computational Linguistics (Volume 1: Long Papers)",
    month = aug,
    year = "2016",
    address = "Berlin, Germany",
    publisher = "Association for Computational Linguistics",
    url = "https://aclanthology.org/P16-1162/",
    doi = "10.18653/v1/P16-1162",
    pages = "1715--1725"
}

@inproceedings{oh-schuler-2024-leading,
    title = "Leading Whitespaces of Language Models' Subword Vocabulary Pose a Confound for Calculating Word Probabilities",
    author = "Oh, Byung-Doh  and
      Schuler, William",
    booktitle = "Proceedings of the 2024 Conference on Empirical Methods in Natural Language Processing",
    month = nov,
    year = "2024",
    address = "Miami, Florida, USA",
    publisher = "Association for Computational Linguistics",
    url = "https://aclanthology.org/2024.emnlp-main.202/",
    doi = "10.18653/v1/2024.emnlp-main.202",
    pages = "3464--3472"
}

@inproceedings{pimentel-meister-2024-compute,
    title = "How to Compute the Probability of a Word",
    author = "Pimentel, Tiago  and
      Meister, Clara",
    editor = "Al-Onaizan, Yaser  and
      Bansal, Mohit  and
      Chen, Yun-Nung",
    booktitle = "Proceedings of the 2024 Conference on Empirical Methods in Natural Language Processing",
    month = nov,
    year = "2024",
    address = "Miami, Florida, USA",
    publisher = "Association for Computational Linguistics",
    url = "https://aclanthology.org/2024.emnlp-main.1020/",
    doi = "10.18653/v1/2024.emnlp-main.1020",
    pages = "18358--18375"
}

@article{schrimpf2021neural,
  title={The neural architecture of language: Integrative modeling converges on predictive processing},
  author={Schrimpf, Martin and Blank, Idan Asher and Tuckute, Greta and Kauf, Carina and Hosseini, Eghbal A. and Kanwisher, Nancy and Tenenbaum, Joshua B. and Fedorenko, Evelina},
  journal={Proceedings of the National Academy of Sciences},
  volume={118},
  number={45},
  pages={e2105646118},
  year={2021},
  publisher={National Acad Sciences},
  url={https://doi.org/10.1073/pnas.2105646118}
}

@article{hosseini2024artificial,
  title={Artificial neural network language models predict human brain responses to language even after a developmentally realistic amount of training},
  author={Hosseini, Eghbal A. and Schrimpf, Martin and Zhang, Yian and Bowman, Samuel and Zaslavsky, Noga and Fedorenko, Evelina},
  journal={Neurobiology of Language},
  volume={5},
  number={1},
  pages={43--63},
  year={2024},
  publisher={MIT Press One Broadway, 12th Floor, Cambridge, Massachusetts 02142, USA~…},
  doi={https://doi.org/10.1162/nol_a_00137}
}

@article{boynton1996linear,
  author = {Boynton, Geoffrey M. and Engel, Stephen A. and Glover, Gary H. and Heeger, David J.},
  title = {Linear Systems Analysis of Functional Magnetic Resonance Imaging in Human V1},
  volume = {16},
  number = {13},
  pages = {4207--4221},
  year = {1996},
  doi = {10.1523/JNEUROSCI.16-13-04207.1996},
  publisher = {Society for Neuroscience},
  URL = {https://www.jneurosci.org/content/16/13/4207},
  eprint = {https://www.jneurosci.org/content/16/13/4207.full.pdf},
  journal = {Journal of Neuroscience}
}

@article{barr2013random,
title = {Random effects structure for confirmatory hypothesis testing: Keep it maximal},
journal = {Journal of Memory and Language},
volume = {68},
number = {3},
pages = {255-278},
year = {2013},
issn = {0749-596X},
doi = {https://doi.org/10.1016/j.jml.2012.11.001},
url = {https://www.sciencedirect.com/science/article/pii/S0749596X12001180},
author = {Dale J. Barr and Roger Levy and Christoph Scheepers and Harry J. Tily}
}

@inproceedings{oh-schuler-2025-impact,
    title = "The Impact of Token Granularity on the Predictive Power of Language Model Surprisal",
    author = "Oh, Byung-Doh  and
      Schuler, William",
    editor = "Che, Wanxiang  and
      Nabende, Joyce  and
      Shutova, Ekaterina  and
      Pilehvar, Mohammad Taher",
    booktitle = "Proceedings of the 63rd Annual Meeting of the Association for Computational Linguistics (Volume 1: Long Papers)",
    month = jul,
    year = "2025",
    address = "Vienna, Austria",
    publisher = "Association for Computational Linguistics",
    url = "https://aclanthology.org/2025.acl-long.209/",
    doi = "10.18653/v1/2025.acl-long.209",
    pages = "4150--4162",
    ISBN = "979-8-89176-251-0"
}

@article{feghhi2024large,
  title={What are large language models mapping to in the brain? a case against over-reliance on brain scores},
  author={Feghhi, Ebrahim and Hadidi, Nima and Song, Bryan and Blank, Idan A and Kao, Jonathan C},
  journal={arXiv preprint arXiv:2406.01538},
  url={https://arxiv.org/abs/2406.01538},
  year={2024}
}

@article{blank2014functional,
  title={A functional dissociation between language and multiple-demand systems revealed in patterns of {BOLD} signal fluctuations},
  author={Blank, Idan and Kanwisher, Nancy and Fedorenko, Evelina},
  journal={Journal of neurophysiology},
  volume={112},
  number={5},
  pages={1105--1118},
  year={2014},
  publisher={American Physiological Society Bethesda, MD},
  url={https://doi.org/10.1152/jn.00884.2013}
}

\appendix
\section{Hyperparameters of Models Examined in this Study}\label{appendix-a}
\begin{table}[h!]
\setlength{\tabcolsep}{4pt}
\centering
\begin{tabular}{lrrrr}
\thickhline
\textbf{Model Variant} & \textbf{\#L} & \textbf{\#H} & \textbf{$d_{model}$} & \textbf{Parameters} \\ \thickhline
GPT-2 Small            & 12           & 12           & 768        & $\sim$124M          \\
GPT-2 Medium           & 24           & 16           & 1024       & $\sim$355M          \\
GPT-2 Large            & 36           & 20           & 1280       & $\sim$774M          \\
GPT2- XL               & 48           & 25           & 1600       & $\sim$1.6B          \\ \hline
GPT-Neo 125M           & 12           & 12           & 768        & $\sim$125M          \\
GPT-Neo 1.3B           & 24           & 16           & 2048       & $\sim$1.3B          \\
GPT-Neo 2.7B           & 32           & 20           & 2560       & $\sim$2.7B          \\
GPT-J 6B               & 28           & 16           & 4096       & $\sim$6B            \\
GPT-NeoX 20B           & 44           & 64           & 6144       & $\sim$20B           \\ \hline
OPT 125M               & 12           & 12           & 768        & $\sim$125M          \\
OPT 350M               & 24           & 16           & 1024       & $\sim$350M          \\
OPT 1.3B               & 24           & 32           & 2048       & $\sim$1.3B          \\
OPT 2.7B               & 32           & 32           & 2560       & $\sim$2.7B          \\
OPT 6.7B               & 32           & 32           & 4096       & $\sim$6.7B          \\
OPT 13B                & 40           & 40           & 5120       & $\sim$13B           \\
OPT 30B                & 48           & 56           & 7168       & $\sim$30B           \\
OPT 66B                & 64           & 72           & 9216       & $\sim$66B           \\ \thickhline
\end{tabular}
\caption{Hyperparameters of the models examined in this study. \#L refers to the number of layers of that model; \#H refers to the number of attention heads each layer has; $d_{model}$ refers to the embedding size of the model.}
\label{table:hyper-para}
\end{table}

\section{Maximal LME models}\label{original-max-lme}

During LME model fitting, we followed \citet{barr2013random} and started with the maximal LME models.
We eliminated one random effect at a time until the LME models converge.
To ensure that the regression includes both by-subject and by-item random effects, we removed random effects iteratively from the by-subject and by-item effects.
That is, if the LME model does not converge, we first removed a random effect from the by-item group; if the LME model still does not converge, we then move on to removing a random effect from the by-subject group, and so on.

Following is the original design of the LME model for Natural Stories fMRI, in which the eliminated random effects are bolded.

\begin{small}
\begin{Verbatim}[commandchars=\\\{\}]
BOLD ~ z.(sentpos) + z.(wlen) + z.(surp) + 
    (1 + \textbf{z.(sentpos)} + z.(wlen) + z.(surp) | subject) + 
    (1 + \textbf{z.(sentpos)} + \textbf{z.(wlen)} + z.(surp) | storyid)
\end{Verbatim}
\end{small}

As for Pereira fMRI, the original design of the LME model is the following, with the eliminated random effect bolded:

\begin{small}
\begin{Verbatim}[commandchars=\\\{\}]
BOLD ~ z.(sentlen) + z.(passpos)+ z.(lastwordsurp) + 
    (1 + z.(sentlen) + z.(passpos) + 
    z.(lastwordsurp) | subject) + (1 + z.(sentlen) + 
    \textbf{z.(passpos)} + z.(lastwordsurp) | sentid)
\end{Verbatim}
\end{small}

\end{document}